%% file: main.tex
\documentclass[10pt,twocolumn,letterpaper]{article}

\usepackage{wacv}              

\input{preamble}

\definecolor{wacvblue}{rgb}{0.21,0.49,0.74}
\usepackage[pagebackref,breaklinks,colorlinks,allcolors=wacvblue]{hyperref}

\def\wacvPaperID{9} 
\def\confName{WACV}
\def\confYear{2026}

\title{RoLID-11K: A Dashcam Dataset for Small-Object Roadside Litter Detection}

\author{Tao Wu$^{1,}$\thanks{Equal contribution.} \quad Qing Xu$^{1,*}$ \quad Xiangjian He$^{1,}$\thanks{Corresponding author.} \quad Oakleigh Weekes$^{2}$ \quad James Brown$^{2}$ \quad Wenting Duan$^{2}$\\
$^1$University of Nottingham Ningbo China \quad $^2$University of Lincoln\\
\tt\small wduan@lincoln.ac.uk}

\begin{document}
\maketitle
\input{0_abstract}

\input{1_introduction}

\input{2_literature}

\input{3_dataset}

\input{4_experiment}

\input{5_discussion}
\input{6_conclusion}
{
    \small
    \bibliographystyle{ieeenat_fullname}
    \bibliography{main}
}

\end{document}

%% file: preamble.tex
\usepackage[T1]{fontenc}


%% file: 0_abstract.tex
\begin{abstract}
Roadside litter poses environmental, safety and economic challenges, yet current monitoring relies on labour-intensive surveys and public reporting, providing limited spatial coverage. Existing vision datasets for litter detection focus on street-level still images, aerial scenes or aquatic environments, and do not reflect the unique characteristics of dashcam footage, where litter appears extremely small, sparse and embedded in cluttered road-verge backgrounds. We introduce RoLID-11K, the first large-scale dataset for roadside litter detection from dashcams, comprising over 11k annotated images spanning diverse UK driving conditions and exhibiting pronounced long-tail and small-object distributions. We benchmark a broad spectrum of modern detectors, from accuracy-oriented transformer architectures to real-time YOLO models, and analyse their strengths and limitations on this challenging task. Our results show that while CO-DETR and related transformers achieve the best localisation accuracy, real-time models remain constrained by coarse feature hierarchies. RoLID-11K establishes a challenging benchmark for extreme small-object detection in dynamic driving scenes and aims to support the development of scalable, low-cost systems for roadside-litter monitoring. The dataset is available at \url{https://github.com/xq141839/RoLID-11K}.
\end{abstract}

%% file: 1_introduction.tex
\section{Introduction}
\label{sec:int}

Litter accumulation along roadsides creates environmental, safety and economic burdens. UK authorities spend hundreds of millions of pounds annually on street cleansing \cite{hm2017litter}, while roadside debris contributes to polluted runoff, obstructed drainage and harm to verge-dwelling wildlife \cite{wu2023applications}. Yet routine monitoring remains inconsistent, typically relying on manual inspections and public reports that provide limited spatial and temporal coverage. The existing commercial litter-detection tools, such as LitterCam and EnviroEye.AI, focus on catching people littering from vehicles via fixed CCTV/pole-mounted cameras, rather than detecting or mapping the accumulation of litter along roadside verges. Moreover, these systems incur substantial deployment and maintenance costs, making large-scale adoption impractical for comprehensive road network coverage.

In contrast, dash cameras are inexpensive, widely used and continuously capture the forward road scene. Their ubiquity in private vehicles and commercial fleets presents a practical opportunity for passive roadside-litter monitoring using video that is already being recorded. However, litter captured from moving vehicles is challenging to detect: objects are typically small, sparse, highly imbalanced, and affected by motion blur, compression and cluttered roadside backgrounds. Existing waste-related datasets such as TACO~\cite{proenca2020taco}, TrashNet~\cite{yang2016trashnet}, UAVVaste~\cite{kraft2021uavlitter} and FloW~\cite{cheng2021flow}, do not reflect these dashcam-specific conditions.

\begin{figure}[!t]
  \centering
  \includegraphics[width=1\linewidth]{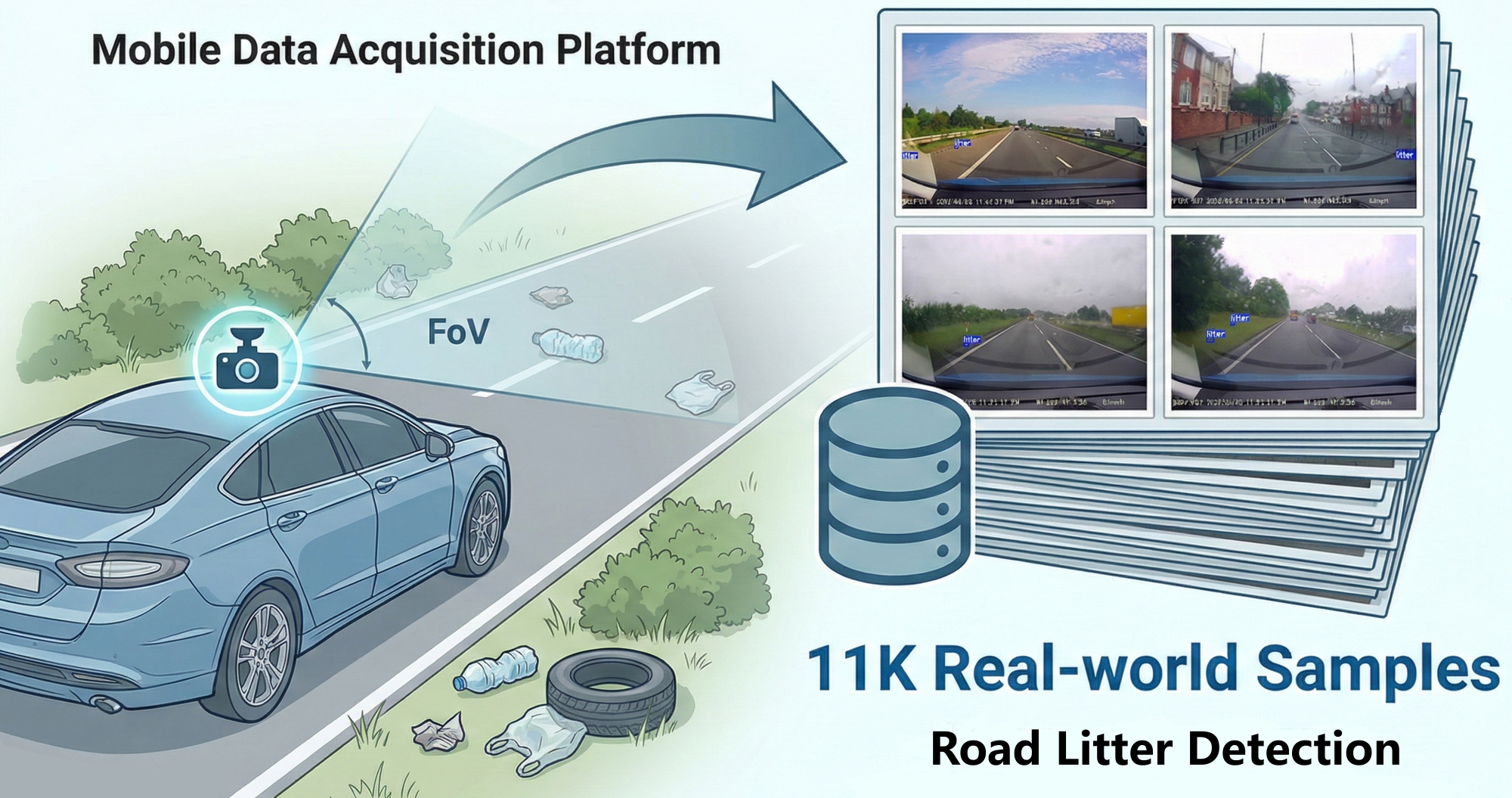}
  \caption{Overview of the proposed RoLID-11K dataset. A vehicle-mounted dashcam serves as a mobile data acquisition platform, capturing roadside litter under diverse real-world driving conditions. The dataset comprises 11K annotated images spanning various weather, lighting, and road environments.}
  \label{fig:intro}
\end{figure}

\begin{figure*}[!t]
  \centering
  \includegraphics[width=1\linewidth]{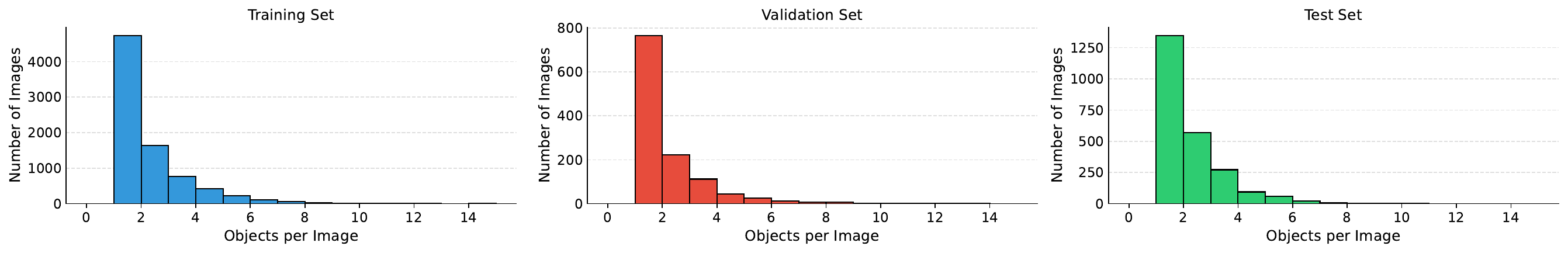}
  \caption{Distribution of object counts per image for training, validation, and test splits in our RoLID-11K dataset. The distribution exhibits a long-tail pattern, reflecting real-world roadside litter scenarios, with most images containing 1–3 objects.}
  \label{fig:1}
\end{figure*}

\begin{figure*}[!t]
  \centering
  \includegraphics[width=1\linewidth]{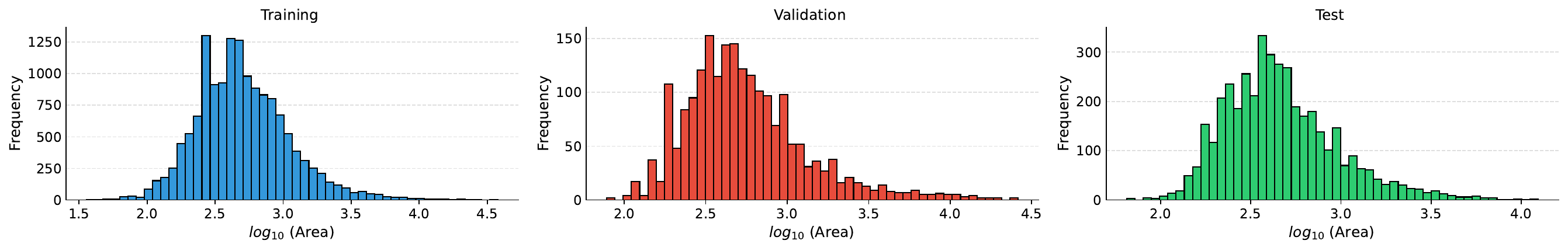}
  \caption{Histogram of object areas in logarithmic scale across dataset splits. The peak around $\log_{10}(\text{Area}) \approx [2.4, 2.8]$ indicates that most litter objects occupy relatively small regions in the image, posing challenges for small object detection.}
  \label{fig:2}
\end{figure*}

\begin{figure*}[!t]
  \centering
  \includegraphics[width=1\linewidth]{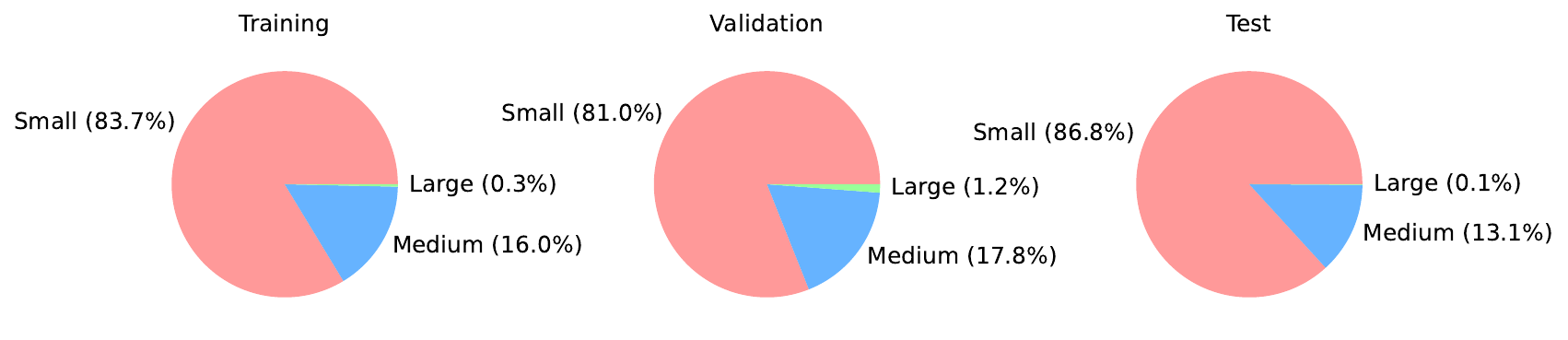}
\caption{Object size distribution following the COCO evaluation criteria: 
small, medium, and large. Small objects dominate across training (83.7\%), validation (81.0\%), and test (86.8\%) splits, underscoring the challenges of small object detection in roadside litter scenarios.}
  \label{fig:3}
\end{figure*}

To address this gap, we introduce RoLID-11K, a Road Litter Detection dataset of over 11,000 annotated dashcam frames featuring real roadside litter with strong long-tail characteristics and a high prevalence of small objects. We benchmark a wide range of modern object detectors, from accuracy-oriented transformer models to real-time YOLO variants, providing the first systematic evaluation of litter detection in dashcam footage. RoLID-11K aims to support scalable, low-cost approaches for tracking roadside pollution and serves as a foundation for future research in small-object detection and dashcam-based environmental monitoring. Our key contributions are:
(i) RoLID-11K, the first large-scale dataset for roadside litter detection from dashcams; 
(ii) an analysis of real-world litter distributions revealing strong long-tail and small-object characteristics; 
(iii) a comprehensive benchmark of state-of-the-art detectors across accuracy and real-time settings; and
(iv) an in-depth insights of benchmark performance, highlighting accuracy–efficiency trade-offs and the challenges posed by dashcam-specific small-object detection.

%% file: 2_literature.tex
\begin{figure*}[!t]
  \centering
  \includegraphics[width=1\linewidth]{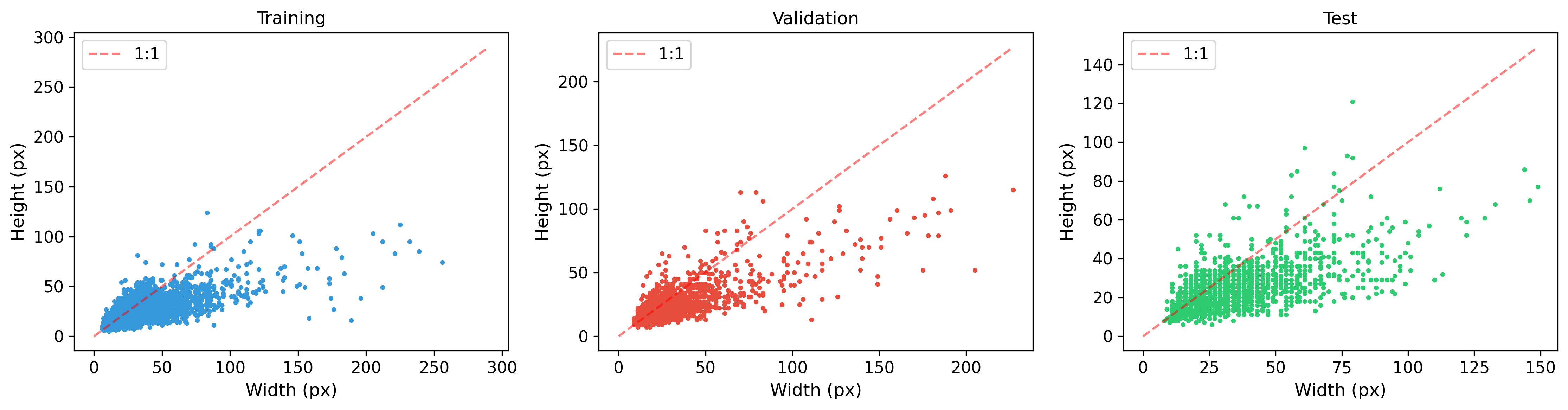}
  \caption{Scatter plot of bounding box dimensions (width vs.\ height) 
across dataset splits. The dashed line indicates a 1:1 aspect ratio. 
Training and validation sets exhibit concentrated distributions with 
similar patterns, while the test set shows more diverse shape variations 
and aspect ratios, providing a challenging benchmark for evaluating 
model robustness and generalization.}
  \label{fig:4}
\end{figure*}

\begin{figure*}[!t]
  \centering
  \includegraphics[width=1\linewidth]{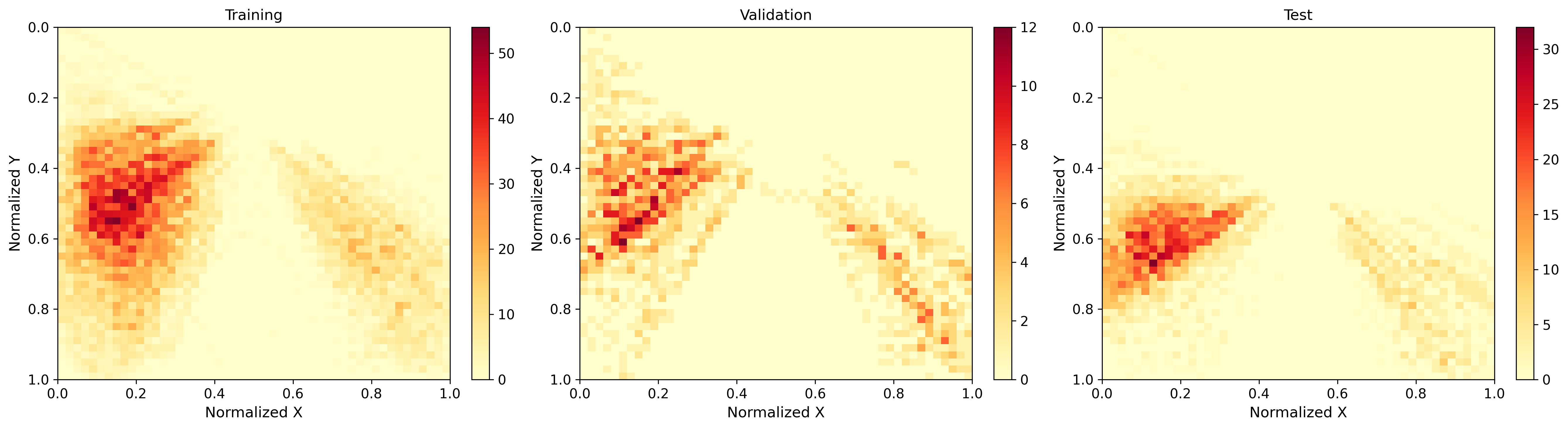}
  \caption{Spatial distribution of object center locations in normalized 
image coordinates. The heatmaps reveal a consistent pattern across 
training, validation, and test sets, with objects concentrated in 
the left-center region corresponding to the roadside area captured 
by vehicle-mounted cameras. This distribution reflects the real-world 
data acquisition setup for roadside litter detection.}
  \label{fig:5}
\end{figure*}

\section{Related Work}
\label{sec:lit}

Existing datasets for litter and waste detection span street-level, aerial and aquatic environments, but none target roadside verges viewed from forward-facing dashcams. At street level, TACO~\cite{proenca2020taco} provides around 1.5k images with multi-class litter annotations, while TrashNet~\cite{yang2016trashnet} offers small-scale single-object classification data. UAV-based datasets such as UAVVaste~\cite{kraft2021uavlitter} supply low-altitude aerial imagery of waste, and several aquatic datasets, e.g., TrashCan~\cite{hong2020trashcan}, TrashICRA~\cite{fulton2019trashicra} and SeaClear~\cite{duras2024seaclear}, focus on marine debris in underwater or surface scenes. PlastOPol~\cite{cordova2022plastopol} also focuses on one-class “litter” detection across diverse outdoor environments using crowdsourced Marine Debris Tracker images, and FloW~\cite{cheng2021flow} targets floating waste in inland waters with both an image-only subset and a multimodal image–radar subset. A recent effort to unify these disparate resources is the DetectWaste benchmark~\cite{majchrowska2022detectwaste}, which standardises annotations across multiple datasets (including extended TACO, UAVVaste, TrashCan, and TrashNet, etc) and corrects label inconsistencies. However, all existing datasets differ markedly from our setting: images are captured from static cameras, handheld devices, UAVs or underwater robots, and litter typically occupies a substantial portion of the frame or appears in clustered patches. None provides large-scale dashcam footage of road verges where litter appears as small, sparse targets along the roadside, which is the focus of RoLID-11K.

Vision-based litter monitoring has been explored in marine, aerial and riverine settings using a range of deep object detectors and segmentors. On underwater datasets such as TrashCan~\cite{hong2020trashcan} and related marine debris collections, baseline experiments typically use Mask R-CNN and Faster R-CNN for instance segmentation and detection, with later work~\cite{lim2023underwaterlitter} comparing lightweight YOLACT to Mask R-CNN for real-time underwater litter segmentation. The SeaClear~\cite{duras2024seaclear} marine debris dataset reports baseline results with Faster R-CNN and YOLOv6, highlighting the difficulty of generalising across sites and cameras. For surface and river-floating waste~\cite{w16101373}, YOLOv5-based pipelines are commonly used for detection and tracking in video streams. Aerial litter detection with UAVVaste~\cite{kraft2021uavlitter} relies on single-stage detectors such as YOLOv4 and EfficientDet deployed on embedded hardware, trading accuracy against onboard inference speed. At a broader level, DetectWaste~\cite{majchrowska2022detectwaste} and recent surveys~\cite{recycling8060086} on automated waste identification show that YOLO variants, together with Faster R-CNN, RetinaNet and related architectures, dominate current waste-detection systems, typically trained on extended TACO and similar datasets. These approaches, however, assume moderate object scales, relatively static viewpoints and domain-specific backgrounds (water surfaces, aerial top-down views, indoor sorting lines), whereas in dashcam footage litter appears as extremely small, sparse targets near the road edge in highly dynamic scenes. This motivates our systematic benchmark of contemporary detectors, covering accuracy-oriented transformers (DINO~\cite{huang2025real}, CO-DETR~\cite{zong2023detrs}, DiffusionDet~\cite{chen2023diffusiondet}) and real-time YOLO models~\cite{cheng2024yolo}, on the new RoLID-11K dataset to characterise their behaviour under combined small-object, long-tail and dashcam-specific challenges.

%% file: 3_dataset.tex
\begin{table*}[!t]
    \centering
    \setlength\tabcolsep{9pt}
    \caption{Comparison of state-of-the-art object detection models on our RoLID-11K dataset.}
    \adjustbox{width=1\textwidth,center}{
    \begin{tabular}{llcccccccc}
    \toprule
    Methods & Publication & Backbone & Epoch & AP$_{50}$ & AP$_{50:95}$ & AP$_{50:95}^{small}$ & AP$_{50:95}^{medium}$ & AP$_{50:95}^{large}$\\
    \midrule
    CO-DETR \cite{zong2023detrs} & ICCV'23 & ResNet-50 & 50 & \textbf{79.2} & \textbf{32.1} & \textbf{31.2} & \textbf{37.5} & \textbf{40.0} \\ 
    DiffusionDet \cite{chen2023diffusiondet} & ICCV'23 & ResNet-50 & 50 & 67.0 & 24.5 & 24.3 & 26.7 & 9.6 \\
    DINO \cite{caron2021emerging} & ICLR'23 & ResNet-50 & 50 & 78.5 & 31.5 & 30.9 & 36.1 & 11.2 \\
    RT-DETR \cite{zhao2024detrs} & CVPR'24 & ResNet-50 & 50 & 73.9 & 28.9 & 28.3 & 32.1 & 18.5 \\
    DEIMv2 \cite{huang2025real} & arXiv'25 & ViT-Tiny & 50 & 74.3 & 27.8 & 27.4 & 30.3 & 21.7 \\
    \bottomrule
    \end{tabular}}
    \label{tab:sota_accuracy}
\end{table*}

\begin{table*}[!t]
    \centering
    \setlength\tabcolsep{8pt}
    \caption{Comparison of real-time object detection models on our RoLID-11K dataset.}
    \adjustbox{width=1\textwidth,center}{
    \begin{tabular}{llcccccccc}
    \toprule
    Methods & Publication & Backbone & Epoch & AP$_{50}$ & AP$_{50:95}$ & AP$_{50:95}^{small}$ & AP$_{50:95}^{medium}$ & AP$_{50:95}^{large}$\\
    \midrule
    YOLOv8 \cite{Jocher_Ultralytics_YOLO_2023} & Ultralytics'23 & CSPDarknet & 50 & 50.1 & 17.5 & 16.6 & 22.9 & 6.0 \\
    YOLOv9 \cite{wang2024yolov9} & ECCV'24 & GELAN & 50 & 50.8 & 17.1 & 16.0 & 23.5 & 4.0 \\
    YOLOv10 \cite{wang2024yolov10} & NeurIPS'24 & CSPDarknet & 50 & 49.7 & 17.4 & 16.3 & 23.2 & 5.1 \\
    YOLOv11 \cite{Jocher_Ultralytics_YOLO_2024} & Ultralytics'24 & C3K2 & 50 & \textbf{52.1} & \textbf{18.3} & \textbf{17.2} & \textbf{24.6} & 5.7 \\
    YOLOv12 \cite{tian2025yolov} & NeurIPS'25 & R-ELAN & 50 & 51.6 & 17.7 & 16.9 & 23.3 & \textbf{15.1} \\
    \bottomrule
    \end{tabular}}
    \label{tab:realtime_accuracy}
\end{table*}

\section{Dataset}
\label{sec:dat}

\subsection{Data Acquisition and Annotation}
RoLID-11K is constructed from 4K dashcam footage recorded in Lincolnshire, UK, between February and July 2022 using a WOLFBOX 4K/1080p dash camera mounted in a standard forward-facing position on a venicle. The videos cover a wide range of driving environments—including rural roads, suburban streets, dual carriageways and urban settings, capturing realistic roadside litter scenarios. They also span diverse weather and lighting conditions, sunny, overcast, low-light and shadowed environments—providing a representative variety of real-world driving scenes. All frames were extracted from the raw videos at their native frame rate using OpenCV’s VideoCapture interface and saved in JPEG Image format, ensuring no content-dependent sampling bias. Frames containing no visible litter were removed to mitigate the substantial natural imbalance between litter and background.

Although the dashcam captures 4K UHD video, extracted frames were standardised to 1920×1080 resolution, downscaling to 1080p preserves the visibility of small litter objects while reducing storage and annotation overhead. During benchmarking, images were further resized according to the input requirements of each detector. All images were anonymised by blurring any visible vehicle number plates and human faces in the frames. Annotations were created using the VGG Image Annotator~\cite{dutta2019via}, with a single class (“litter”) and bounding boxes drawn around any visible item of litter, including very small or partially occluded objects embedded in vegetation. This yielded 14,645 annotated instances in the training set, 2,094 in the validation set, and 4,189 in the test set.

\begin{table}[!t]
    \centering
    \setlength\tabcolsep{3pt}
    \caption{Model complexity and inference speed comparison.}
    \adjustbox{width=0.48\textwidth,center}{\begin{tabular}{lcccc}
    \toprule
    Methods & Image Size & \#Param(M) & FLOPs(G) & Latency(ms)\\
    \midrule
    CO-DETR \cite{zong2023detrs} & 800$\times$1333 & 64.5 & 267.5 & 6.0 \\ 
    DiffusionDet \cite{chen2023diffusiondet} & 800$\times$1333 & 72.3 & 192.8 & 24.3 \\
    DINO \cite{caron2021emerging} & 800$\times$1333 & 47.5 & 274.0 & 6.6 \\
    RT-DETR \cite{zhao2024detrs} & 640$\times$640 & 32.0 & 103.4 & 2.8 \\
    DEIMv2 \cite{huang2025real} & 640$\times$640 & 9.7 & 25.4 & 10.7 \\
    \midrule
    YOLOv8 \cite{Jocher_Ultralytics_YOLO_2023} & 640$\times$640 & 3.0 & 8.1 & 0.8 \\
    YOLOv9 \cite{wang2024yolov9} & 640$\times$640 & \textbf{2.0} & 7.6 & 1.0 \\
    YOLOv10 \cite{wang2024yolov10} & 640$\times$640 & 2.3 & 6.5 & \textbf{0.6} \\
    YOLOv11 \cite{Jocher_Ultralytics_YOLO_2024} & 640$\times$640 & 2.6 & \textbf{6.3} & 0.8 \\
    YOLOv12 \cite{tian2025yolov} & 640$\times$640 & 2.6 & 6.3 & 0.9 \\
    \bottomrule
    \end{tabular}}
    \label{tab:complexity}
\end{table}

\begin{table*}[!t]
    \centering
    \setlength\tabcolsep{6pt}
    \caption{Impact of backbone on object detection methods with different architectures.}
    \adjustbox{width=1\textwidth,center}{
    \begin{tabular}{lccccccccccccc}
    \toprule
    Methods & Backbone & AP$_{50}$ & AP$_{50:95}$ & AP$_{50:95}^{ small}$ & AP$_{50:95}^{medium}$ & AP$_{50:95}^{large}$ & \#Param(M) & FLOPs(G) & Latency(ms)\\
    \midrule
    \multirow{3}{*}{DEIMv2 \cite{huang2025real}} & HGNetv2 & 71.8 & 26.1 & 25.8 & 28.4 & 10.7 & 3.5 & 6.8 & 8.8 \\ 
    & ViT-Tiny & 74.3 & 27.8 & 27.4 & 30.3 & 21.7 & 9.7 & 25.4 & 10.2 \\
     & ViT-Tiny+ & 74.2 & 27.3 & 26.5 & 30.6 & 10.5 & 18.0 & 51.9 & 10.4 \\
    \midrule
    & R-ELAN-N & 51.6 & 17.7 & 16.9 & 23.3 & 15.1 & 2.6 & 6.3 & 0.9 \\ 
    & R-ELAN-S & 55.7 & 20.3 & 19.2 & 26.1 & 13.3 & 9.2 & 21.2 & 1.1 \\
    YOLOv12 \cite{tian2025yolov} & R-ELAN-M & 56.7 & 20.8 & 19.9 & 27.2 & 5.6 & 20.1 & 67.1 & 1.48 \\
    & R-ELAN-L & 56.8 & 21.0 & 19.9 & 27.6 & 18.1 & 26.3 & 88.5 & 1.98 \\
    & R-ELAN-X & 55.5 & 20.6 & 19.4 & 26.8 & 14.6 & 59.0 & 198.5 & 2.73 \\
    \bottomrule
    \end{tabular}}
    \label{tab1}
\end{table*}

\subsection{Dataset Split and Statistics}
The final dataset consists of 11,565 images, divided into 7990 training, 1201 validation, and 2374 test images, i.e., the splits used in our benchmark. RoLID-11K exhibits several challenging characteristics for object detection. The number of objects per image follows a strong long-tail distribution, with most images containing one to three instances (Figure~\ref{fig:1}). Object sizes are extremely small: distributions of bounding-box areas peak around $\log_{10}(\text{Area})$ $\approx$ 2.4–2.8, meaning that litter typically occupies only a tiny portion of each frame as shown in Figure~\ref{fig:2}. According to COCO size criteria, over 80\% of all annotated objects are classified as small across all splits (Figure~\ref{fig:3}). Bounding-box aspect-ratio analysis further shows high variability, with the test set in particular exhibiting diverse object shapes as illustrated by Figure ~\ref{fig:4}, increasing the difficulty of robust detection. Finally, Figure ~\ref{fig:5} shows the object-centre heatmaps, revealing a strong spatial bias toward the lower-left region of the image, reflecting typical UK left-hand driving where the dashcam predominantly captures the left road verge. Litter also tends to accumulate on this side due to driver behaviour (discarding items toward the verge) and wind-driven displacement, making the left verge more frequently populated than the right. Together, these properties make RoLID-11K a demanding benchmark for evaluating small-object detection under real-world driving conditions.

%% file: 4_experiment.tex
\section{Experiments}
\label{sec:exp}

\subsection{Benchmark Design and Rationale}
RoLID-11K represents an extremely challenging setting for object detection due to its high proportion of very small objects, strong long-tail instance distribution, and dynamic dashcam viewpoint. To meaningfully evaluate performance under these conditions, we benchmark two complementary families of detectors:
\begin{itemize}
    \item \textbf{Accuracy-oriented transformer architectures} (CO-DETR~\cite{zong2023detrs}, DiffusionDet~\cite{chen2023diffusiondet}, DINO~\cite{caron2021emerging}, RT-DETR~\cite{zhao2024detrs}, and DEIMv2~\cite{huang2025real}), which are known to excel in localisation precision and small-object sensitivity on datasets such as COCO~\cite{lin2014microsoft}.
    \item \textbf{Real-time architectures} (YOLOv8~\cite{Jocher_Ultralytics_YOLO_2023} - YOLOv12~\cite{tian2025yolov}), widely used in automotive and edge applications where inference speed is crucial.
\end{itemize}
This combination allows us to assess the trade-off between accuracy and deployability, and to identify which architectural trends, like transformer-based modelling, dense prediction heads, or real-time convolutions, are most effective for dashcam-based litter detection. We include RT-DETR and DEIMv2 as modern attempts to bridge high accuracy and real-time inference, and multiple YOLO generations to reflect the practical relevance of lightweight detectors in real-time roadside monitoring systems. This selection covers the current spectrum of contemporary detectors (2021–2025), ensuring that our benchmark reflects the state of the art.

\subsection{Implementation Details}
We perform all experiments on a workstation equipped with an Intel Xeon Silver 4216 CPU, 256GB RAM, and an NVIDIA H200 GPU (141GB memory). Models are trained with their framework-provided defaults to ensure comparability and reproducibility. Transformers (CO-DETR, DiffusionDet, DINO) are implemented using MMDetection~\cite{chen2019mmdetection}, while YOLO-series models, RT-DETR and DEIMv2 use the Ultralytics framework~\cite{Jocher_Ultralytics_YOLO_2023}. For MMDetection-based detectors, the input resolution is set to $800 \times 1333$ following the standard COCO protocol. For YOLO-series, RT-DETR and DEIMv2, we use the default input size of $640 \times 640$. All models are initialized with weights pre-trained on COCO~\cite{lin2014microsoft} and fine-tuned on our training set for 50 epochs. Inference latency is measured with batch size 1 over the full test set, using averaged runtime in milliseconds per frame. These measures (as shown in Table~\ref{tab:complexity}), allow direct comparison of accuracy–efficiency trade-offs.

\subsection{Evaluation Metrics}

We adopt the standard COCO evaluation protocol \cite{lin2014microsoft} to comprehensively assess detection performance on our RoLID-11K dataset. The primary metrics include Average Precision (AP). Specifically. we report AP$_{50}$, which measures detection accuracy at an IoU threshold of 0.5, and AP$_{50:95}$, which averages AP across IoU thresholds from 0.5 to 0.95 with a step of 0.05. AP$_{50:95}$ provides a more stringent evaluation of localization quality. Moreover, given the prevalence of small objects in roadside litter scenarios, we report AP$_{50:95}^{small}$ for small objects (area $< 32^2$ px$^2$), AP$_{50:95}^{medium}$ for medium objects ($32^2 \leq$ area $< 96^2$ px$^2$), and AP$_{50:95}^{large}$ for large objects (area $\geq 96^2$ px$^2$). These metrics are particularly important for evaluating model performance on the challenging small object detection task inherent to our dataset. To assess computational efficiency for practical deployment, we report the number of parameters (\#Param), floating-point operations (FLOPs), and inference latency measured in milliseconds per image.

\begin{figure*}[!t]
  \centering
  \includegraphics[width=0.83\linewidth]{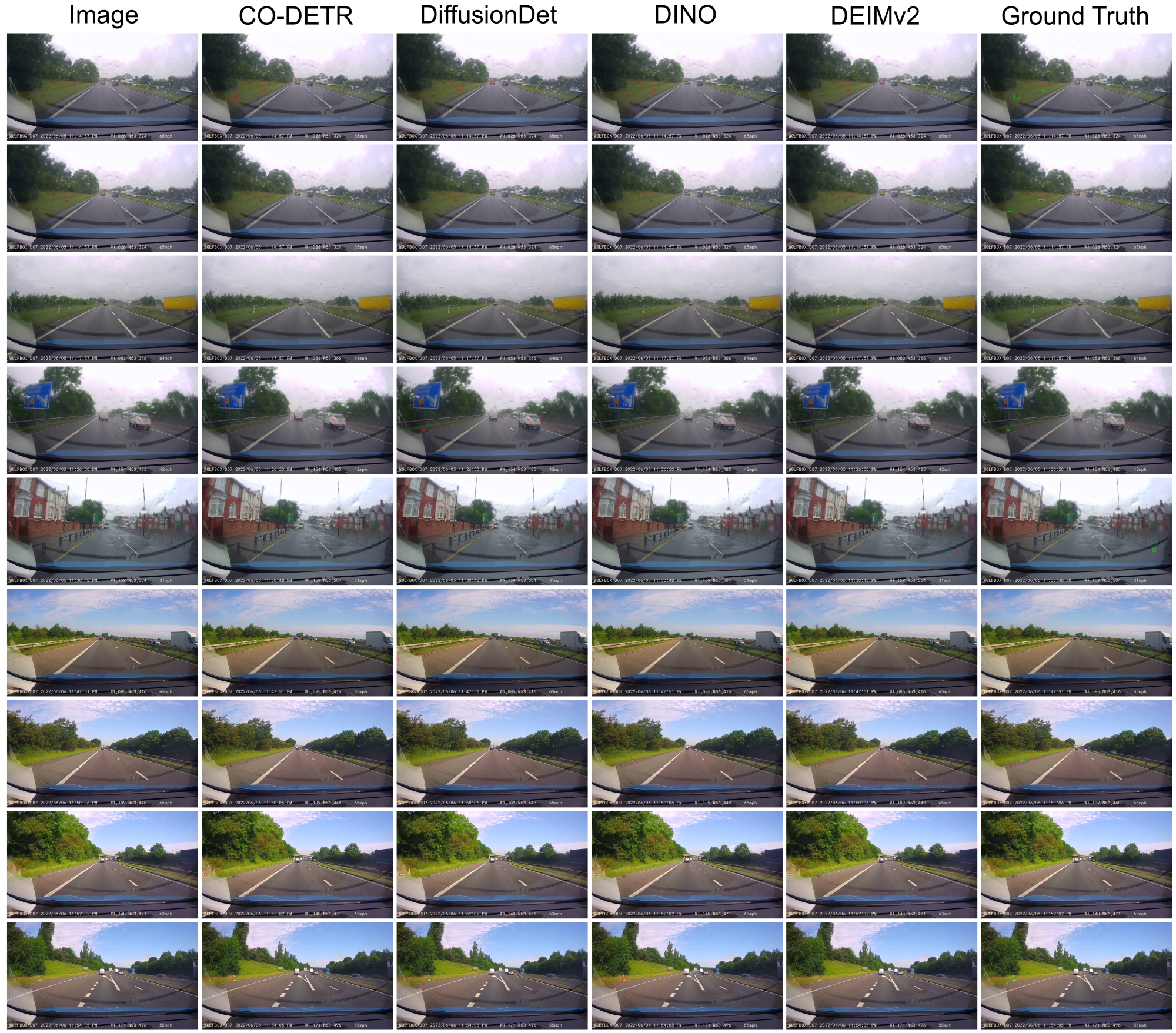}
\caption{Qualitative comparison of state-of-the-art detectors on the RoLID-11K test set with the MMDetection platform.}
  \label{fig:6}
\end{figure*}

\begin{figure*}[!t]
  \centering
  \includegraphics[width=0.95\linewidth]{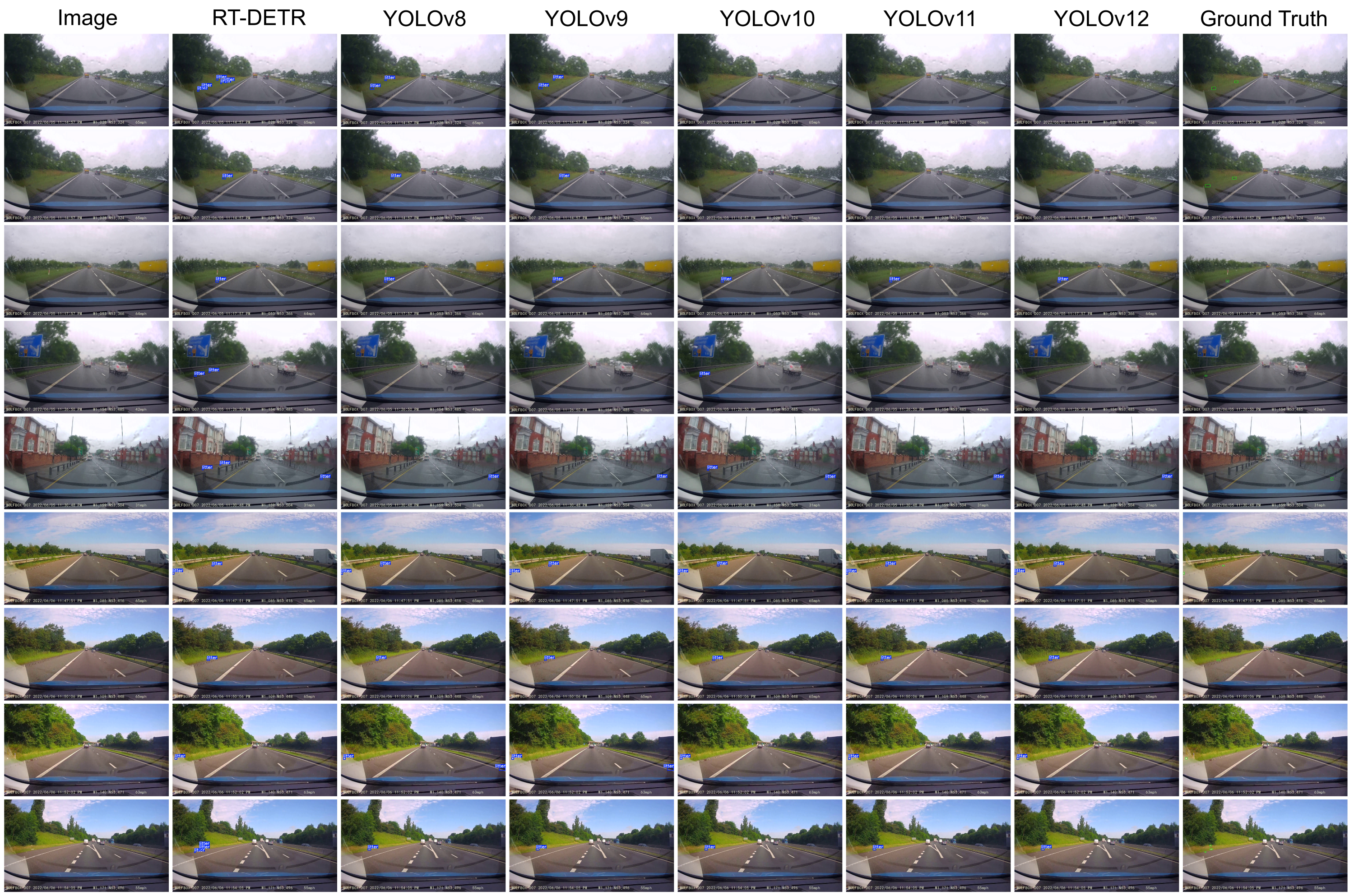}
\caption{Qualitative comparison of state-of-the-art detectors on the RoLID-11K test set with the Ultralytics platform.}
  \label{fig:7}
\end{figure*}

%% file: 5_discussion.tex
\section{Results and Discussion}
\label{sec:dis}

Tables~\ref{tab:sota_accuracy} and ~\ref{tab:realtime_accuracy} summarise the performance of accuracy-oriented and real-time detectors. Among transformer-based models, CO-DETR achieves the highest AP$_{50:95}$ confirming its strong localisation ability and robustness to the extreme small-object distribution characteristic of RoLID-11K. DINO also performs competitively, whereas DiffusionDet underperforms on this dataset, suggesting that its coarse denoising schedule struggles with detecting tiny objects embedded in cluttered backgrounds. The generally higher AP$_{50}$ relative to AP$_{50:95}$ across models reflects the substantial challenge of precise localisation for objects occupying only a few dozen pixels. 

Real-time detectors exhibit the expected trade-off between speed and accuracy. YOLO models (v8–v12) achieve sub-millisecond inference latency while maintaining competitive AP$_{50}$ scores but lag significantly in AP$_{50:95}$ compared with transformer architectures. This performance gap is most pronounced for AP$_{50:95}^{small}$, reinforcing that lightweight detection heads and lower input resolution limit fine-grained localisation on very small targets.

Table~\ref{tab1} shows that backbone choice has a marked impact on performance. We ablate DEIMv2 and YOLOv12 as they are the most recent representatives of their respective model families and offer modular backbones that make architectural comparisons meaningful. For DEIMv2, replacing the default CNN backbone with ViT-Tiny yields a noticeable improvement in AP$_{50:95}$. This aligns with the observation that transformer-based encoders preserve long-range contextual information and fine spatial detail, which is crucial for detecting litter objects measuring only a few pixels. For YOLOv12, improvements in backbone design and prediction heads yield modest gains in, though all versions remain limited by input resolution and lightweight feature hierarchies when detecting very small litter objects. These results suggest that architectural capacity in the early feature extraction stages is a key factor for small-object detection under the RoLID-11K conditions.

Figures~\ref{fig:6} and ~\ref{fig:7} illustrate model predictions on challenging scenes. Accuracy-oriented detectors capture small, partially occluded items more reliably, whereas real-time models frequently miss objects embedded in vegetation or shadowed regions. YOLO variants tend to produce more false negatives but maintain stable detections on medium-sized objects when present. Transformer models reduce false negatives but occasionally produce false positives on textured roadside regions, reflecting the cluttered background typical of dashcam imagery. These examples highlight the difficulty of balancing precision and recall when objects are both visually subtle and spatially biased toward the image boundaries.

Overall, our results show that among all evaluated models, CO-DETR achieves the strongest overall AP$_{50:95}$, indicating that dense transformer-based assignment mechanisms provide the most reliable localisation for extremely small and sparsely distributed litter instances. However, while accuracy-oriented transformer detectors achieve the best performance, their computational cost limits real-time deployment on low-power platforms. Conversely, YOLO models provide extremely fast inference but struggle to capture the fine spatial detail required for consistent small-object detection. This tension underscores the need for architectures specifically tailored to extreme small-object regimes, potentially combining high-resolution feature pathways with efficient inference mechanisms. The RoLID-11K benchmark exposes these limitations clearly and provides a basis for future work on models capable of operating effectively in real-time roadside monitoring.

%% file: 6_conclusion.tex
\section{Conclusion}
\label{sec:con}

We introduced RoLID-11K, the first large-scale dataset for roadside litter detection from dashcam video, capturing the challenges of real-world monitoring where objects are extremely small, sparse and spatially biased toward road verges. Through a benchmark of contemporary detectors, we showed that accuracy-oriented transformer architectures currently provide the strongest localisation performance, while real-time YOLO models, despite their speed, struggle with the fine spatial detail required for detecting litter-sized objects. These findings highlight the need for architectures specifically tailored to extreme small-object detection in dynamic driving environments. RoLID-11K establishes a foundation for deployable models for environmental monitoring, and we hope it will support the development of low-cost systems for tracking roadside pollution.

\section*{Acknowledgements}

This work is partially supported by the Yongjiang Technology Innovation Project (2022A-097-G), Zhejiang Department of Transportation General Research and Development Project (2024039), and National Natural Science Foundation of China grant (62476037).